\documentclass[10pt,conference]{IEEEtran}
\IEEEoverridecommandlockouts

\newif\ifieee
\ieeefalse

\usepackage{multirow}
\usepackage{booktabs}
\usepackage{float}
\usepackage{amsmath,amssymb,amsfonts}
\usepackage{algorithmic}
\usepackage{graphicx}
\usepackage{textcomp}
\usepackage[dvipsnames,table,xcdraw]{xcolor}
\def\BibTeX{{\rm B\kern-.05em{\sc i\kern-.025em b}\kern-.08em
    T\kern-.1667em\lower.7ex\hbox{E}\kern-.125emX}}
\usepackage{soul} %
\setuldepth{Berlin}
\usepackage{xspace}

\usepackage[caption=false,farskip=0pt]{subfig}
\usepackage[hyphens]{url}
\ifieee
\else
\usepackage[colorlinks=true,allcolors=black]{hyperref}
\fi
\usepackage[acronym]{glossaries}
\usepackage[sort,compress]{cite} %
\usepackage{balance}
\usepackage{arydshln}
\usepackage[T1]{fontenc}
\usepackage[capitalise]{cleveref} %

\newacronymstyle{long-short-br}
{%
  \GlsUseAcrEntryDispStyle{long-short}%
}%
{%
  \GlsUseAcrStyleDefs{long-short}%
}
\setacronymstyle{long-short-br}

\usepackage{transparent}
\usepackage{tikz}
\newcommand\copyrighttext{%
  \scriptsize Accepted at SIBGRAPI 2024. The final published version is available on IEEE Xplore (DOI: \href{https://doi.org/10.1109/SIBGRAPI62404.2024.10716303}{\textcolor{blue}{10.1109/SIBGRAPI62404.2024.10716303}}).}
\newcommand\copyrightnotice{%
\begin{tikzpicture}[remember picture,overlay]
\node[anchor=south,yshift=30pt,xshift=0pt] at (current page.south) {\fbox{\transparent{0.85}\parbox{\dimexpr0.77\textwidth-\fboxsep-\fboxrule\relax}{\copyrighttext}}};
\end{tikzpicture}%
}

\newif\ifarxiv
\arxivfalse

\newif\iffinal
\finaltrue

\newcommand{\cmtid}{55}
\newcommand{\papertitle}{Enhancing License Plate Super-Resolution: A~Layout-Aware and Character-Driven~Approach} 
\iffinal
\else
\usepackage[switch]{lineno}
\fi

\begin{document}

\iffinal

\title{\papertitle}

\author{
\begin{tabular}{cc}
\multicolumn{2}{c}{
\hspace{1.5mm}Valfride Nascimento\IEEEauthorrefmark{1}, Rayson Laroca\IEEEauthorrefmark{2}$^,$\IEEEauthorrefmark{1}, Rafael O. Ribeiro\IEEEauthorrefmark{3}, William Robson Schwartz\IEEEauthorrefmark{4}, David Menotti\IEEEauthorrefmark{1}} \\[0.5ex] \hspace{8.5mm}\small
\IEEEauthorrefmark{1}\hspace{0.15mm}Federal University of Paran\'a, Curitiba, Brazil & \small \IEEEauthorrefmark{2}\hspace{0.15mm}Pontifical Catholic University of Paran\'a, Curitiba, Brazil \\[-2.5pt]
\hspace{8.5mm}\small \IEEEauthorrefmark{3}\hspace{0.15mm}Brazilian Federal Police, Bras\'{\i}lia, Brazil & \small \IEEEauthorrefmark{4}\hspace{0.15mm}Federal University of Minas Gerais, Belo Horizonte, Brazil\ \\
\multicolumn{2}{c}{\resizebox{0.9\linewidth}{!}{
\hspace{1.5mm}\IEEEauthorrefmark{1}{\hspace{-0.45mm}\tt\normalsize \{vwnascimento,menotti\}@inf.ufpr.br} \; \IEEEauthorrefmark{2}{\hspace{0.05mm}\tt\normalsize rayson@ppgia.pucpr.br} \; \IEEEauthorrefmark{3}{\tt\normalsize rafael.ror@pf.gov.br}  \; \IEEEauthorrefmark{4}\hspace{0.2mm}{\tt\normalsize william@dcc.ufmg.br}
}}
\end{tabular}
}

\else
    \title{\papertitle}

    \author{SIBGRAPI paper ID: \cmtid \\[8ex]}
    \linenumbers
\fi

\newcommand*{\DM}[2][]{\textcolor{red}{[\textbf{\ifthenelse{\equal{#1}{}}{DM}{DM(#1)}}: #2]}}
\newcommand*{\VN}[2][]{\textcolor{blue}{[\textbf{\ifthenelse{\equal{#1}{}}{VN}{VN(#1)}}: #2]}}
\newcommand*{\RL}[2][]{\textcolor{Rhodamine}{[\textbf{\ifthenelse{\equal{#1}{}}{RL}{RL(#1)}}: #2]}}

\maketitle
\ifieee
    {\let\thefootnote\relax\footnote{\\979-8-3503-7603-6/24/\$31.00
    \textcopyright2024 IEEE}}
\else
    \copyrightnotice
\fi

\newcommand*{\todo}[2][]{\textcolor{red}{[\textbf{\ifthenelse{\equal{#1}{}}{TODO}{TODO(#1)}}: #2]}}

\newcommand\red[1]{{\textcolor{red}{#1}}}
\newcommand\orange[1]{{\textcolor{BurntOrange}{\textbf{#1}}}}

\newcommand\review[1]{{#1}} %

\newacronym{cnn}{CNN}{Convolutional Neural Network}
\newacronym{hr}{HR}{high-resolution}
\newacronym{lp}{LP}{License Plate}
\newacronym{lpr}{LPR}{License Plate Recognition}
\newacronym{lr}{LR}{low-resolution}
\newacronym{misr}{MISR}{Multi-Image Super-Resolution}
\newacronym{mse}{MSE}{Mean Squared Error}
\newacronym{ocr}{OCR}{Optical Character Recognition}
\newacronym{psnr}{PSNR}{Peak Signal-to-Noise Ratio}
\newacronym{sisr}{SISR}{Single-Image Super-Resolution}
\newacronym{srcnn}{SRCNN}{Super-Resolution Convolutional Neural Network}
\newacronym{ssim}{SSIM}{Structural Similarity Index Measure}
\newacronym{vsr}{VSR}{Video Super-Resolution}
\newacronym{rcb}{RCB}{Residual Concatenation Block}
\newacronym{fm}{FM}{Feature Module}
\newacronym{sfe}{SFE}{Shallow Feature Extractor}
\newacronym{rm}{RM}{Reconstruction Module}
\newacronym{ps}{PS}{\textit{PixelShuffle}}
\newacronym{pu}{PU}{\textit{PixelUnshuffle}}
\newacronym{nn}{NN}{Neural Network}
\newacronym{psfe}{PSFE}{Pre-shallow Feature Extractor}
\newacronym{alpr}{ALPR}{Automatic License Plate Recognition}
\newacronym{ca}{CA}{Channel Unit}
\newacronym{pos}{POS}{Positional Unit}
\newacronym{tfam}{TFAM}{Two-fold Attention Module}
\newacronym{map}{MAP}{Maximum a Posteriori}
\newacronym{gan}{GAN}{Generative Adversarial Network}
\newacronym{ccpd}{CCPD}{Chinese City Parking Dataset}
\newacronym{mprnet}{MPRNet}{Multi-Path Residual Network}
\newacronym{cbam}{CBAM}{Convolution Block Attention Module}
\newacronym{se}{SE}{Squeeze-and-excitation}
\newacronym{esa}{ESA}{Enhanced Spatial Attention}
\newacronym{csrgan}{CSRGAN}{Character-Based Super-Resolution Generative Adversarial Networks}
\newacronym{dconv}{DConv}{depthwise-separable convolutional layer}
\newacronym{gp}{GP}{Geometrical Perception Unit}
\newacronym{srgan}{SRGAN}{Super-Resolution Generative Adversarial Networks}
\newacronym{dpca}{DPCA}{Dual-Coordinate Direction Perception Attention}
\newacronym{dganesr}{D\textunderscore GAN\textunderscore ESR}{Double Generative Adversarial Networks for Image Enhancement and Super Resolution}
\newacronym{lcofl}{LCOFL}{Layout and Character Oriented Focal Loss}
\newacronym{gplpr}{GP\_LPR}{Global Perception License Plate Recognition}

\newacronym{rdb}{RDB}{Residual Dense Block}
\newacronym{pltfam}{PLTFAM}{Pixel Level Three-Fold Attention Module}
\newacronym{lpd}{LPD}{License Plate Detection}
\newacronym{plnet}{PLNET}{Pixel-Level Network}

\newacronym{gt}{GT}{Ground Truth}
\newcommand{\rodosolalpr}{RodoSol-ALPR\xspace}
\newcommand{\ufpralpr}{UFPR-ALPR\xspace}

\iffinal
\newcommand{\supplementary}{\url{https://github.com/valfride/lpsr-lacd}}

\else

\newcommand{\supplementary}{\textit{[hidden for review]}}

\fi
\ifieee
\vspace{-3.575mm}
\else
\vspace{-3.575mm}
\fi
\begin{abstract}

Despite significant advancements in \gls*{lpr} through deep learning, most improvements rely on high-resolution images with clear characters.
This scenario does not reflect real-world conditions where traffic surveillance often captures low-resolution and blurry images.
Under these conditions, characters tend to blend with the background or neighboring characters, making accurate \gls*{lpr} challenging.
To address this issue, we introduce a novel loss function, \gls*{lcofl}, which considers factors such as resolution, texture, and structural details, as well as the performance of the \gls*{lpr} task itself.
We enhance character feature learning using deformable convolutions and shared weights in an attention module and employ a GAN-based training approach with an \gls*{ocr} model as the discriminator to guide the super-resolution process.
Our experimental results show significant improvements in character reconstruction quality, outperforming two state-of-the-art methods in both quantitative and qualitative measures.
Our code is publicly available at \supplementary.

\end{abstract}

\glsresetall
\section{Introduction}

\gls*{alpr} systems have become increasingly popular in recent years, driven by their diverse practical applications, including toll collection, traffic monitoring, and forensic investigations~\cite{laroca2021efficient,moussa2022forensic,liu2024irregular}.

These systems generally encompass two main tasks: \gls*{lpd} and \gls*{lpr}.
\gls*{lpd} is concerned with locating the areas in an image that contain \glspl*{lp}, whereas \gls*{lpr} is dedicated to identifying the characters on these \glspl*{lp}.
Recent studies have been particularly concentrated on the \gls*{lpr} stage.
Although the reported recognition rates have typically been high, most research has been conducted on high-resolution \glspl*{lp}, where the characters are easily discernible and clearly defined~\cite{silva2022flexible,laroca2023leveraging,rao2024license}.

Surveillance cameras typically record images with low resolution or poor quality~\cite{santos2022face}, mainly because of bandwidth and storage constraints.
This leads to a scenario where \gls*{lp} characters blend with the background and adjacent characters, seriously affecting the effectiveness of \gls*{lpr} systems~\cite{nascimento2022combining,nascimento2023super,pan2024lpsrgan}.

Various image enhancement techniques, including super-resolution and denoising methods, have been developed to improve image quality by increasing resolution or overall clarity.
Despite significant strides in this field, most approaches focus on enhancing objective image quality metrics such as \gls*{psnr} or \gls*{ssim} without considering the specific application at hand~\cite{mehri2021mprnet,liu2023blind}.
These techniques often struggle to differentiate between similar characters in \gls*{lr} images, such as `B' and `8', `G' and `6', and `T' and~`7'.

In this work, we propose a perceptual loss function named \textit{\gls*{lcofl}} to enhance \gls*{lp} super-resolution and, consequently, \gls*{lpr} performance.
\gls*{lcofl} guides the network's learning by considering not only factors such as resolution, texture, and structural details, but also the performance of the \gls*{lpr} task itself.
It specifically penalizes errors related to character confusion and layout inconsistencies (i.e.,~\gls*{lp} sequences that deviate from the patterns observed in the training set), thus mitigating incorrect reconstructions.
To further improve performance, we incorporate a \gls*{gan}-style training, leveraging predictions from an \gls*{ocr} model as the discriminator.
Notably, \gls*{lcofl} can utilize predictions from any \gls*{ocr} model since it relies solely on raw character~predictions.

In summary, the main contributions of this work are:

\begin{itemize}
    \item A novel loss function crafted to elevate the reconstruction of \gls*{lp} characters by integrating character recognition within the super-resolution process;
    \item Improvements to prevalent architectures in prior works by incorporating deformable convolution layers and shared weights into the attention module.
    A GAN-based training approach is also proposed, employing an \gls*{ocr} model as the discriminator.
    These strategies are aimed at generating \gls*{lp} images that are not only of high quality and resolution but also more accurately recognizable by \gls*{ocr}~models;
    \item We have made our source code publicly available, aiming to stimulate further research within this~domain.
\end{itemize}

\section{Related Work}
\label{sec:related_work}

In this section, we review related works.
First, we discuss approaches used for \gls*{lpr} in \cref{RelatedWork:LPR}.
Then, we cover super-resolution methods for \gls*{lpr} in \cref{RelatedWork:SRLP}.

\subsection{License Plate Recognition}
\label{RelatedWork:LPR}

Silva \& Jung~\cite{silva2020realtime} proposed to frame \gls*{lpr} as an object detection task with individual characters treated as unique classes for detection and recognition.
They developed CR-NET, a YOLO-based model tailored specifically for \gls*{lpr}, which has subsequently proven highly effective and has been adopted in several follow-up studies~\cite{laroca2021efficient,oliveira2021vehicle,silva2022flexible}.

Recent advancements have brought forth segmentation-free approaches using \glspl*{lp} to classify text from an entire LP image as a sequence.
Ke et al.~\cite{ke2023ultra} developed a lightweight multi-scale LPR network with depthwise separable convolutional residual blocks and a multi-scale feature fusion layer for quick inference.
Liu et al.~\cite{liu2024improving} introduced deformable spatial attention modules to integrate global layout, enhancing character feature extraction.
Rao et al.~\cite{rao2024license} tackled large-angle LP deflections caused by cameras by integrating channel attention mechanisms into CRNN, thereby improving \gls*{lpr}~performance.

Multi-task models have also demonstrated considerable success for \gls*{lpr}.
In these models, convolutional layers first process the entire \gls*{lp} image.
Then, the network splits into $N$ separate branches with fully connected layers.
Each branch is dedicated to recognizing a single character (or blank), enabling the simultaneous prediction of up to $N$ characters~\cite{goncalves2018realtime,fan2022improving,wang2022rethinking}.

While these methods have shown considerable success in \gls*{lpr}, most of them were trained and evaluated using \gls*{hr} images. 
However, this scenario does not mirror actual surveillance conditions, where \gls*{lr} images are prevalent.
\gls*{lr} images often have characters that blend into the \gls*{lp} background due to quality issues and compression techniques used for storage~\cite{maier2022reliability,moussa2022forensic,nascimento2023super}.

\subsection{Super-Resolution for License Plate Recognition}
\label{RelatedWork:SRLP}

Recent advancements in deep learning have significantly improved general text super-resolution.
However, there remains a notable gap in research dedicated to specifically enhancing \gls*{lp} recognition in low-resolution~images.

Lin et al.~\cite{lin2021license} proposed using SRGAN~\cite{ledig2017photo} for \gls*{lp} super-resolution.
Their approach was subsequently refined by Hamdi et al.~\cite{hamdi2021new}, who introduced Double Generative Adversarial Networks~(D\_GAN\_ESR\_) for denoising, deblurring, and super-resolving \glspl*{lp}.
However, these approaches lacked integration of character recognition within the learning process.

Focusing more on \gls*{lp} recognition, Pan et al.~\cite{pan2023super} proposed a pipeline for LP super-resolution and recognition, exploring ESRGAN~\cite{wang2019esrgan} for single-character super-resolution.
While effective, this method struggles with heavily degraded \glspl*{lp} where character boundaries are unclear.
To enhance that approach, the same authors~\cite{pan2024lpsrgan} later introduced a new degradation model to simulate low-resolution \glspl*{lp} and proposed LPSRGAN, an extension of ESRGAN that processes entire \gls*{lp} images.
LPSRGAN, trained with an \gls*{ocr}-based loss function, better preserves character details and \gls*{lp} structural features but generates incorrect text in complex degradation cases~\cite{pan2024lpsrgan}.

Kim et al.~\cite{kim2024afanet} developed AFA-Net, combining super-resolution with pixel and feature-level deblurring to address motion blur in \glspl*{lp}.
Although their findings showed potential, their test methodology was somewhat oversimplified, featuring \gls*{lr} images that remain legible and focusing solely on super-resolving digits while excluding letters and Korean~characters.

Lee et al.~\cite{lee2020super} devised a GAN-based model incorporating a character perceptual loss utilizing features from ASTER, a well-known scene text recognition model.
More recently, Nascimento et al.~\cite{nascimento2023super} introduced subpixel-convolution layers and an attention module to enhance textural and structural details, also with a perceptual loss integrating an \gls{ocr} model.
Despite their achievements, both methods encountered difficulties related to character confusion caused by structural or font similarities, resulting in reconstructions that deviated from the expected patterns in the specific \gls*{lp} layout under~ investigation.

\section{Proposed Approach}
\label{sec:proposed}

This section details the proposed methodology.
In \cref{proposedApproach:lcofl}, we introduce our novel perceptual loss function, focusing on enhancing character reconstruction while accounting for the \gls*{lp} layout.
In \cref{proposedApproach:am}, we outline the model's architecture, expanding on the framework introduced in~\cite{nascimento2023super}, where it achieved significant success.

\subsection{Layout and Character Oriented Focal Loss}
\label{proposedApproach:lcofl}

\glsreset{lcofl}

In low-resolution \glspl*{lp}, characters often lose their shape details, blending with the \gls*{lp}'s background and neighboring characters.
Moreover, the lack of precise character positioning relative to the \gls*{lp} layout often causes the model to incorrectly super-resolve a letter as a digit or vice versa~\cite{lorch2019forensic, nascimento2022combining, nascimento2023super}.

To enhance network guidance for \gls*{lp} reconstruction, we designed the \gls*{lcofl} based on four key insights:
(i) treating reconstruction partially as a classification task, where the super-resolved characters within an \gls*{lp} image need to be correctly identified by an \gls*{ocr} model;
(ii) recognizing that characters typically adhere to specific patterns based on the \gls*{lp} layout (this includes fixed positions for digits and letters), which should be maintained during reconstruction;
(iii) ensuring accurate reconstruction of characters with similar structures (e.g., ``2'' and ``Z'' or ``R'' and ``B'');
and (iv) preserving structural details from the original \gls*{gt} images in the final super-resolved~images.

\subsubsection{Classification Loss}
To tackle the \gls*{lp} recognition problem, we adopt a weighted cross-entropy loss within the \gls*{lcofl} function, as defined in \cref{cofl:ce}:
\begin{equation}
L_C = -\frac{1}{K}\sum_{k=1}^{K}w_k\log p_t (y^{GT}_{k} | x_k) \, ,
\label{cofl:ce}
\end{equation}

\noindent where $p_t(y^{GT}_{k})$ represents the predicted probabilities of the encoded \gls*{gt} label~$y^{GT}_t$ for the $k$-th character, while $x$ represents the predicted probabilities for the super-resolved images.
$K$ is the maximum decoding length for the \gls*{ocr}~alphabet.

The weights $w$ are initialized as $[1, \dots, K]$, where no penalization is applied, and each position corresponds to an encoded character in the \gls*{ocr} alphabet.
These weights act as penalties assigned to characters misclassified due to structural or font similarities.
Following the validation phase in each epoch, a confusion matrix is generated to assess the recognition of super-resolved images by the \gls*{ocr} model against the \gls*{gt} labels.
Based on the confusion matrix, pairs of characters frequently confused with each other are identified, and an $\alpha$ value is added to the respective position of the encoded character in $w$, which is utilized during the training~phase.

\subsubsection{LP Layout Penalty}

In various regions, including the one explored in this study (Brazil), digits and letters adhere to a fixed positional arrangement within the \glspl*{lp}.
This means that a digit should not be mistakenly reconstructed as a letter, and vice versa.
To enforce this constraint, a layout penalty, as defined in \cref{cofl:lp}, is incorporated into the overall loss function.
\begin{equation}  
    \begin{split}
        L_{P} = \sum_{i=1}^{K} \left[D(x_k) \cdot A(y_{k}^{GT}) + \cdot A(x_k) \cdot D(y_{k}^{GT}) \right]
    \end{split}
    \label{cofl:lp}
\end{equation}

In \cref{cofl:lp}, $D(\cdot)$ indicates the assertion of a digit at a specific position, while $A(\cdot)$ denotes the assertion of a letter.
The left side of the sum verifies the correct positioning of a letter, while the right side checks the placement of a digit.
As per the criteria outlined in \cref{eq:cases}, there is no penalty added if a character is correctly positioned.
However, for each misplacement, a penalty value~$\beta$ is added to the sum. 
\begin{equation}
    \begin{aligned}
        \text{D}(c) &= \begin{cases}
            \beta & \text{if } c \text{ is a digit} \\
            0 & \text{otherwise}
        \end{cases} \\
        \text{A}(c) &= \begin{cases}
            \beta & \text{if } c \text{ is a letter} \\
            0 & \text{otherwise}
        \end{cases}
    \end{aligned}
    \label{eq:cases}
\end{equation}

\subsubsection{Dissimilarity Loss}

Most objective methods for assessing image quality compare a reference, distortion-free image with a sample image.
One commonly used metric is \gls*{mse}, which measures the average squared difference between each pixel of the reference and the sample image.
\gls*{mse} is also used in \gls*{psnr} to calculate the ratio between the maximum possible value of a signal (the reference image) and its corrupted counterpart.
These metrics are popular due to their simplicity and clear physical meaning~\cite{jamil2024review}.

However, they do not align well with the human visual system, which excels at identifying structural aspects of a scene, such as contrast, textures, structures, and luminance differences \cite{cheon2021ambiguity}.
To better guide the network in super-resolving images with a focus on structural information, we incorporate \gls*{ssim}~\cite{wang2004image} into our loss function, as shown in \cref{cofl:ssim}:
\begin{equation}
    \label{cofl:ssim}
    L_{S} = \frac{1 - SSIM(S_i, H_i)}{2} \, ,
\end{equation}

\noindent where $S_ i$ represents the super-resolved image, and $H_ i$ stands for the high-resolution \gls*{gt} image.
\gls*{ssim} evaluates three key aspects: luminance, contrast, and structure.
\gls*{ssim} values range from -$1$ to $1$, with values near -$1$ indicating very different images and values near $1$ indicating highly similar images in terms of structural~similarity.

In \cref{cofl:ssim}, the transformation $(1-SSIM)/2$ adjusts the \gls*{ssim} values to a range of $[0, 1]$.
This adjustment ensures that $0$ represents highly similar images, while $1$ represents highly dissimilar images.
This adjustment facilitates its integration into the loss function, enabling more effective penalization of the network for generating images that deviate significantly from the \gls*{gt} in terms of structure.

After considering the classification loss~($L_C$), the \gls*{lp} layout penalty~($L_{P}$), and the dissimilarity loss~($L_{S}$), the final loss function is formulated in \cref{cofl:final}.
The main goal is to improve the recognition rates attained by an \gls*{ocr} model when dealing with \gls*{lr} images.
Simultaneously, it aims to restore intricate details to facilitate subsequent forensic analysis, which often represents the ultimate goal of the super-resolution~process.
\begin{equation}
    \label{cofl:final}
    loss = L_{D} + L_{P} + L_{S}
\end{equation}

\subsection{Architecture}
\label{proposedApproach:am}

We extend the network architecture introduced in~\cite{nascimento2023super}, specifically addressing the arrangement of \glspl*{pltfam} within the~\glspl*{rcb}~\cite{mehri2021mprnet}.

Nascimento et al.~\cite{nascimento2023super} focused on the design of the \gls*{pltfam}.
Essentially, this design strategy capitalizes on \textit{PixelShuffle}'s capabilities to utilize inter-channel feature relationships within the Channel Unit.
It also integrates positional localization through the Positional Unit and improves texture and reconstruction with the Geometrical Perception Unit.
Despite achieving remarkable recognition results, the attention module is uniquely defined in each \gls*{rcb}, which hampers the attention module's learning potential across individual~\glspl*{rcb}.

To overcome this limitation, we propose incorporating an attention module with shared weights throughout the network structure.
This enables the attention module to gather information from the initial layers, which extract fundamental \gls*{lp} characteristics, to the final layers, where finer details are restored.
In this way, the attention module can consistently emphasize learning features essential for accurate \gls*{lp}~recognition.

Furthermore, we have replaced the depthwise convolutions with deformable convolution layers in both the Positional and Channel units.
Traditional convolution layers apply fixed geometric transformations, which may not adequately capture intricate, non-uniform deformations present in \gls*{lr} or degraded \gls*{lp} images.
This limitation can restrict the network's capability to handle the diverse spatial arrangements of characters effectively.
In contrast, deformable convolution layers dynamically adjust their receptive fields based on input features, allowing for more adaptable and precise modeling of spatial dependencies~\cite{liu2024irregular}.
This adaptability significantly enhances the network's capacity to accurately reconstruct characters for recognition, even under challenging conditions.
Consequently, the model's overall \gls*{lpr} performance is improved~\cite{zhu2019deformable, liu2024irregular}.
\section{Experiments}
\label{sec:experiments}

This section covers our experiments, starting with the experimental setup.
We then analyze the results, emphasizing significant improvements in recognition accuracy and reconstruction quality.
Afterward, we conduct an ablation study to evaluate each component's contribution to the overall results.
\review{Finally, we present initial experiments conducted on real-world~data.}

\subsection{Setup}
\label{setup}

\begin{figure}[!htb]
    \centering
    
    \resizebox{0.925\linewidth}{!}{
    \includegraphics[width=0.48\linewidth, height=0.16\linewidth]{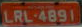} \hspace{-0.7mm}
    \includegraphics[width=0.48\linewidth, height=0.16\linewidth]{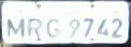} \hspace{-0.7mm}
    \includegraphics[width=0.48\linewidth, height=0.16\linewidth]{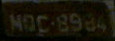} \hspace{-0.7mm}
    \includegraphics[width=0.48\linewidth, height=0.16\linewidth]{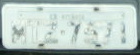}
    }
    
    \vspace{0.4mm}
    
    \resizebox{0.925\linewidth}{!}{
    \includegraphics[width=0.48\linewidth, height=0.16\linewidth]{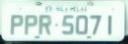} \hspace{-0.7mm}
    \includegraphics[width=0.48\linewidth, height=0.16\linewidth]{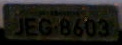} \hspace{-0.7mm}
    \includegraphics[width=0.48\linewidth, height=0.16\linewidth]{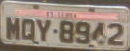} \hspace{-0.7mm}
    \includegraphics[width=0.48\linewidth, height=0.16\linewidth]{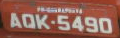}
    }
    
    \vspace{0.4mm}
    
    \resizebox{0.925\linewidth}{!}{
    \includegraphics[width=0.48\linewidth, height=0.16\linewidth]{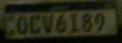} \hspace{-0.7mm}
    \includegraphics[width=0.48\linewidth, height=0.16\linewidth]{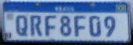} \hspace{-0.7mm}
    \includegraphics[width=0.48\linewidth, height=0.16\linewidth]{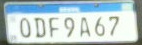} \hspace{-0.7mm}
    \includegraphics[width=0.48\linewidth, height=0.16\linewidth]{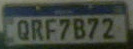}
    }
    
    \vspace{0.4mm}
    
    \resizebox{0.925\linewidth}{!}{
    \includegraphics[width=0.48\linewidth, height=0.16\linewidth]{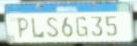} \hspace{-0.7mm}
    \includegraphics[width=0.48\linewidth, height=0.16\linewidth]{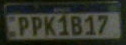} \hspace{-0.7mm}
    \includegraphics[width=0.48\linewidth, height=0.16\linewidth]{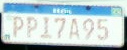} \hspace{-0.7mm}
    \includegraphics[width=0.48\linewidth, height=0.16\linewidth]{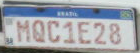}
    }
    
    \vspace{-2mm}
    
    \caption{Some \gls*{lp} images from the \rodosolalpr dataset~\cite{laroca2022cross}. The first two rows show Brazilian \glspl*{lp}, while the last two show Mercosur \glspl*{lp}. This work focuses on \glspl*{lp} with all characters arranged in a single row (10k~images).}
    \label{fig:samples-rodosol}
\end{figure}

We conducted our experiments using the \rodosolalpr dataset~\cite{laroca2022cross}, which includes $10{,}000$ images of cars obtained at toll stations.
This dataset includes $5{,}000$ images featuring cars with Brazilian \glspl*{lp} and another $5{,}000$ depicting cars with Mercosur \glspl*{lp}\footnote{Following prior literature~\cite{laroca2022first,nascimento2022combining,silva2022flexible}, we use the term ``Brazilian'' to refer to the \gls*{lp} layout used in Brazil prior to the adoption of the Mercosur~layout.}.
Brazilian \glspl*{lp} are composed of $3$ letters followed by $4$ digits, while the initial pattern adopted for Mercosur \glspl*{lp} in Brazil comprises $3$ letters, $1$ digit, $1$ letter, and $2$ digits.
We chose the \rodosolalpr dataset due to its diversity and frequent use in recent research~\cite{nascimento2022combining,laroca2023leveraging,liu2024irregular}.
In \cref{fig:samples-rodosol}, we present \glspl*{lp} extracted from the dataset (cropped and rectified), showcasing differences in lighting, color combinations, and character~fonts.

We explored the \gls*{lr}-\gls*{hr} pairs created and made available by Nascimento et al.~\cite{nascimento2023super}.
The \gls*{lr} versions of each \gls*{hr} image were generated by simulating the effects of a low-resolution optical system with heavy degradation due to environmental factors or compression techniques. This was done by iteratively applying random Gaussian noise and resizing with bicubic interpolation until reaching the desired degradation level of SSIM~$<0.1$.

\begin{figure}[!htb]
    \centering
    
    \resizebox{0.925\linewidth}{!}{
    \includegraphics[width=0.54\linewidth, height=0.18\linewidth]{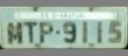} \hspace{-0.5mm}
    \includegraphics[width=0.54\linewidth, height=0.18\linewidth]{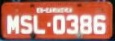} \hspace{-0.5mm}
    \includegraphics[width=0.54\linewidth, height=0.18\linewidth]{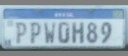} \hspace{-0.5mm}
    \includegraphics[width=0.54\linewidth, height=0.18\linewidth]{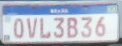} 
    }
    
    \vspace{0.6mm}
    
    \resizebox{0.925\linewidth}{!}{
    \includegraphics[width=0.54\linewidth, height=0.18\linewidth]{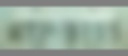} \hspace{-0.5mm}
    \includegraphics[width=0.54\linewidth, height=0.18\linewidth]{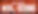} \hspace{-0.5mm}
    \includegraphics[width=0.54\linewidth, height=0.18\linewidth]{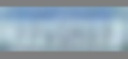} \hspace{-0.5mm}
    \includegraphics[width=0.54\linewidth, height=0.18\linewidth]{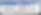}
    }    
    
    \vspace{-2.5mm}
    \caption{Examples of \gls*{hr}-\gls*{lr} image pairs used in our~experiments.}
    \label{fig:SSIM_examples}
\end{figure}

We also applied padding to the \gls*{lr} and \gls*{hr} images using gray pixels to preserve their aspect ratio before resizing them to $16\times48$ and $32\times96$ pixels, respectively, which corresponds to an upscale factor of~$2$.
\cref{fig:SSIM_examples} shows examples of \gls*{lp} images obtained through this~process.

In our GAN-based training methodology, we employed the \gls*{ocr} model proposed by Liu et al.~\cite{liu2024irregular}~(GP\_LPR) as the discriminator and the super-resolution model proposed in~\cite{nascimento2023super}~(\gls*{plnet}) as the~generator.
During the testing phase, we employed the multi-task \gls*{ocr} model proposed by Gon{\c{c}}alves et al.~\cite{goncalves2018realtime}, which has demonstrated significant success in previous studies~\cite{nascimento2022combining,nascimento2023super}.
The decision to use different \gls*{ocr} models during training and testing was made to prevent biased reconstructions in the testing~process.

The Adam optimizer was used with a learning rate of $10$\textsuperscript{-$4$} for all models.
To address oscillations during training, we implemented the StepLR learning rate schedule as recommended by Liu et al.~\cite{liu2024irregular}.
This approach entails reducing the learning rate by a factor of $0.9$ every $5$ epochs if no improvement in recognition rate is observed.
The GP\_LPR model was set up with a decoding length of $K=7$, corresponding to the 7-character format found in \glspl*{lp} in the \rodosolalpr~dataset.

The experiments were performed using PyTorch framework on a computer with an AMD Ryzen $9$ $5950$X~CPU, $128$~GB of RAM, and an NVIDIA Quadro RTX $8000$ GPU ($48$~GB).

\subsection{Experimental Results}

\gls*{lpr} models are typically evaluated using recognition rates, defined as the number of correctly recognized \glspl*{lp} divided by the total number of \glspl*{lp} in the test set~\cite{silva2022flexible,wang2022rethinking,ke2023ultra}.
Given the prevalence of low-quality/low-resolution images in surveillance systems, we also report partial matches where at least six or at least five characters are correctly recognized.
These partial matches are valuable in forensic applications because they can significantly narrow down the list of candidate~\glspl*{lp}.

The results are presented in \cref{tab:ExperimentalResults}. 
The first two rows show the recognition rates achieved by the \gls*{ocr} model~\cite{goncalves2018realtime} on the original \gls*{hr} images and their corresponding \gls*{lr} counterparts. 
This performance serves as a baseline to demonstrate the challenges faced by \gls*{ocr} models in low-resolution scenarios. 
As can be seen, the \gls*{ocr} model performs poorly on low-resolution \glspl*{lp}, achieving only a $1.1\%$ recognition~rate.

\begin{table}[!htb]
\centering
\setlength{\tabcolsep}{5pt}
\caption{Recognition rates achieved on different test images.}

\vspace{-2mm}

\resizebox{0.725\linewidth}{!}{
\begin{tabular}{lccc}
\toprule
\multirow{2}{*}{Test Images} & \multicolumn{3}{c}{\# Correct Characters} \\[-1pt] \cmidrule(l{11pt}r{9pt}){2-4} \\[-10pt]
                              & All          & $\geq6$          & $\geq5$          \\
\midrule                
\gls*{hr} (original images) & 98.5\% & 99.9\% & 99.9\% \\
\gls*{lr} (degraded images)    &  \phantom{0}1.1\% & \phantom{0}5.3\% & 14.3\% \\
\midrule
LR + SR (\gls*{plnet}~\cite{nascimento2023super}) &  39.0\% & 59.9\% & 74.2\% \\
LR + SR (SR3~\cite{saharia2023image})  & 43.1\% & 67.5\% & 82.2\% \\
LR + SR (\textbf{Proposed})  & \textbf{49.8\%} & \textbf{71.2\%} & \textbf{83.3\%} \\

\bottomrule
\end{tabular} \,
}
\label{tab:ExperimentalResults}
\end{table}

In the lower section of \cref{tab:ExperimentalResults}, we present the \gls*{ocr} model's performance on \gls*{lr} images enhanced with super-resolution techniques.
Our super-resolution network considerably outperformed two state-of-the-art baselines~\cite{nascimento2023super,saharia2023image}.
Specifically, the \gls*{ocr} model showed notable performance improvements when the \glspl*{lp} were super-resolved by our model.
It attains a recognition rate of $49.8$\%, in contrast to $43.1$\% using SR3~\cite{saharia2023image}, a renowned diffusion method by Google Research, and $39.0$\% by PLNET~\cite{nascimento2023super}.
Notably, our model took approximately $38$ minutes to super-resolve all $4$k test images (batch size~=~$1$), whereas SR3 required around $33$ hours ($52\times$~slower), thus underscoring our method's superior accuracy coupled with significantly reduced computational processing time.
\review{Naturally, to ensure a fair comparison, all models (i.e., our proposed approach and the baselines) were trained on the same samples.}

\begin{figure}[!htb]
    \captionsetup[subfigure]{labelformat=empty,position=top,captionskip=0.75pt,justification=centering}
    
    \centering
    \resizebox{1\linewidth}{!}{
    \subfloat[LR (Input)]{
    \includegraphics[width=0.21\linewidth, height=0.064\linewidth]{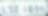}
    } \hspace{-2.75mm}
    \subfloat[\gls*{plnet}~\cite{nascimento2023super}]{
    \includegraphics[width=0.21\linewidth, height=0.064\linewidth]{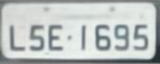}
    } \hspace{-2.75mm}
    \subfloat[SR3~\cite{saharia2023image}]{    
    \includegraphics[width=0.21\linewidth, height=0.064\linewidth]{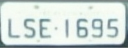}
    } \hspace{-2.75mm}
    \subfloat[Proposed]{
    \includegraphics[width=0.21\linewidth, height=0.064\linewidth]{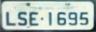}
    } \hspace{-2.75mm}
    \subfloat[HR (GT)]{
    \includegraphics[width=0.21\linewidth, height=0.064\linewidth]{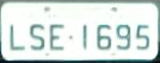} 
    } \,
    }

    \vspace{-2.3mm}

    \resizebox{1\linewidth}{!}{
    \subfloat[]{
    \includegraphics[width=0.21\linewidth, height=0.064\linewidth]{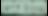}
    } \hspace{-2.75mm}
    \subfloat[]{
    \includegraphics[width=0.21\linewidth, height=0.064\linewidth]{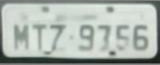}
    } \hspace{-2.75mm}
    \subfloat[]{
    \includegraphics[width=0.21\linewidth, height=0.064\linewidth]{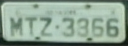}
    } \hspace{-2.75mm}
    \subfloat[]{
    \includegraphics[width=0.21\linewidth, height=0.064\linewidth]{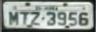}
    } \hspace{-2.75mm}
    \subfloat[]{
    \includegraphics[width=0.21\linewidth, height=0.064\linewidth]{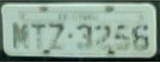} 
    } \,
    } 

    \vspace{-2.3mm}

    \resizebox{1\linewidth}{!}{
    \subfloat[]{
    \includegraphics[width=0.21\linewidth, height=0.064\linewidth]{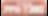}
    } \hspace{-2.75mm}
    \subfloat[]{
    \includegraphics[width=0.21\linewidth, height=0.064\linewidth]{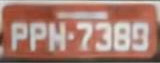}
    } \hspace{-2.75mm}
    \subfloat[]{
    \includegraphics[width=0.21\linewidth, height=0.064\linewidth]{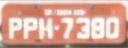}
    }\hspace{-2.75mm}
    \subfloat[]{
    \includegraphics[width=0.21\linewidth, height=0.064\linewidth]{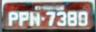}
    } \hspace{-2.75mm}
    \subfloat[]{
    \includegraphics[width=0.21\linewidth, height=0.064\linewidth]{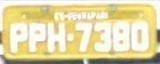} 
    } \,
    }
    
    \vspace{-2.3mm}
     
    \resizebox{1\linewidth}{!}{
    \subfloat[]{
    \includegraphics[width=0.21\linewidth, height=0.064\linewidth]{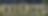}
    } \hspace{-2.75mm}
    \subfloat[]{
    \includegraphics[width=0.21\linewidth, height=0.064\linewidth]{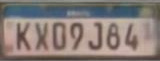}
    } \hspace{-2.75mm}
    \subfloat[]{
    \includegraphics[width=0.21\linewidth, height=0.064\linewidth]{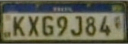}
    } \hspace{-2.75mm}
    \subfloat[]{
    \includegraphics[width=0.21\linewidth, height=0.064\linewidth]{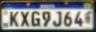}
    } \hspace{-2.75mm}
    \subfloat[]{
    \includegraphics[width=0.21\linewidth, height=0.064\linewidth]{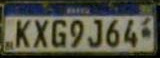} 
    } \,
    } 
    
    \vspace{-2.3mm}

    \resizebox{1\linewidth}{!}{
    \subfloat[]{
    \includegraphics[width=0.21\linewidth, height=0.064\linewidth]{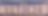}
    } \hspace{-2.75mm}
    \subfloat[]{
    \includegraphics[width=0.21\linewidth, height=0.064\linewidth]{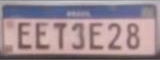} 
    } \hspace{-2.75mm}
    \subfloat[]{
    \includegraphics[width=0.21\linewidth, height=0.064\linewidth]{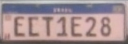} 
    } \hspace{-2.75mm}
    \subfloat[]{
    \includegraphics[width=0.21\linewidth, height=0.064\linewidth]{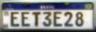} 
    } \hspace{-2.75mm}
    \subfloat[]{
    \includegraphics[width=0.21\linewidth, height=0.064\linewidth]{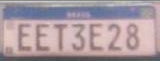} 
    } \,
    }
    
    \vspace{-2.3mm}

    \resizebox{1\linewidth}{!}{
    \subfloat[]{
    \includegraphics[width=0.21\linewidth, height=0.064\linewidth]{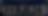}
    } \hspace{-2.75mm}
    \subfloat[]{
    \includegraphics[width=0.21\linewidth, height=0.064\linewidth]{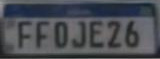}
    } \hspace{-2.75mm}
    \subfloat[]{
    \includegraphics[width=0.21\linewidth, height=0.064\linewidth]{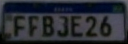}
    } \hspace{-2.75mm}
    \subfloat[]{
    \includegraphics[width=0.21\linewidth, height=0.064\linewidth]{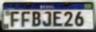}
    } \hspace{-2.75mm}
    \subfloat[]{
    \includegraphics[width=0.21\linewidth, height=0.064\linewidth]{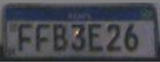} 
    } \,
    }

    \caption{Representative images produced by the proposed approach and baseline methods for the same inputs. GT = Ground Truth.}
    \label{fig:Qresults}
\end{figure}

In \cref{fig:Qresults}, we present a comparison of images enhanced by our method with those using the baselines~\cite{nascimento2023super,saharia2023image}.
Remarkably, the proposed approach outperforms the baseline methods by accurately distinguishing between letters and digits, maintaining the original texture and structural details, and achieving superior visual quality in the~reconstruction.

\Gls*{plnet}~\cite{nascimento2023super} performs quite well in reconstructing textures and structures but faces challenges with character confusion.
For instance, it mistakenly reconstructed an ``S'' as a ``5'' in the top row and a ``G'' as a ``0'' in the fourth row.
In contrast, the SR3 model~\cite{saharia2023image} accurately reconstructs characters according to the \gls*{lp} layout, reducing confusion between letters and digits.
Nevertheless, SR3 still presents inconsistencies, such as partially reconstructing the letter ``E'' without its central line in the fifth row and exhibiting variations in the curvature of the two ``F'' in the bottom~row.

The proposed approach showcases superior character reconstruction by consistently generating well-defined, super-resolved characters that adhere closely to the \gls*{lp} layout.
This results in minimal discrepancies, particularly those arising from poor lighting conditions.
For instance, in the final row of \cref{fig:Qresults}, while both baseline methods reconstructed a ``J'' due to a light spot, our approach aimed for a ``3,'' aligning more accurately with the digit's structure. 
Furthermore, our method excels in maintaining consistent shapes and contours of characters across various LPs, leading to more reliable reconstructions.
This is evident when analyzing the curves of the letter ``E'' and the two ``F''s in the bottom two~rows.

\subsection{Ablation Study}

This work introduces a novel loss function and incorporates several modifications into the training procedure, including a GAN-based style and changes to architectures proposed in previous works, particularly in~\cite{nascimento2023super}.
To assess the contribution of each component, this section presents an ablation~study.

We trained several networks under various conditions.
These conditions included implementations with and without the proposed architectural modifications~(ArchMod) detailed in \cref{proposedApproach:am}, with and without the GAN-based training style~(GAN-style) described in \cref{setup}, and utilizing either our proposed loss function~(\gls*{lcofl}) or the loss function from Nascimento et al.\cite{nascimento2023super}, which incorporates the logits from the \gls*{ocr} model proposed by Goncalves et al.~\cite{goncalves2018realtime}.
In the experiments without the GAN-based style, predictions from the OCR model~\cite{goncalves2018realtime} on the super-resolved images served as input for the LCOFL loss (similar to~\cite{nascimento2023super}).
These experiments led to seven baselines, and the results are detailed in \cref{tab:ab-table}.

\begin{table}[!htb]
\centering
\caption{Recognition rates~(RR) achieved with different components integrated into the proposed approach.}
\label{tab:ab-table}

\vspace{-2mm}

\resizebox{0.825\linewidth}{!}{
\begin{tabular}{lc}
\toprule
Approach &  RR \\ \midrule
Proposed (w/o ArchMod, GAN-style, and LCOFL) & $39.0$\% \\
Proposed (w/o LCOFL)                         & $45.9$\% \\ 
Proposed (w/o ArchMod and LCOFL)             & $47.6$\% \\
Proposed (w/o GAN-style and LCOFL)           & $47.7$\% \\
Proposed (w/o ArchMod and GAN-style)                       & $48.2$\% \\
Proposed (w/o ArchMod)         & $49.2$\% \\
Proposed (w/o GAN-style)                     & $49.4$\% \\ 
\midrule
\textbf{Proposed}                                     & $\textbf{49.8}$\textbf{\%} \\ \bottomrule

\end{tabular} \,
}
\end{table}

The ablation results demonstrate that each component contributes to the overall performance.
Excluding the \gls*{lcofl} leads to a significant performance decrease (from $49.8$\% to $45.9$\%).
Removing ArchMod and GAN-style training also results in performance drops, albeit less severe, to $49.2$\% and $49.4$\%, respectively.
Notably, excluding all the proposed components reduces the recognition rate drastically to just~$39.0$\%.

Based on these results, we argue that \gls*{lcofl} plays a crucial role in aiding the network to accurately position the characters according to the \gls*{lp} layout and mitigate potential confusion caused by structural or font similarities among characters.
Additionally, incorporating shared weights in PLTFAM and integrating deformable convolutions into its architecture have significantly enhanced the attention module's ability to extract structural and textural features relevant to the~characters.

\subsection{\review{Preliminary Experiments on Real-World Data}}

\begin{figure}[!htb]
    \centering
    
    \includegraphics[width=0.46\linewidth]{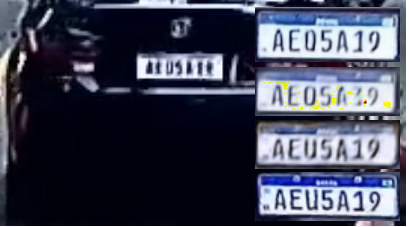}
    \includegraphics[width=0.46\linewidth]{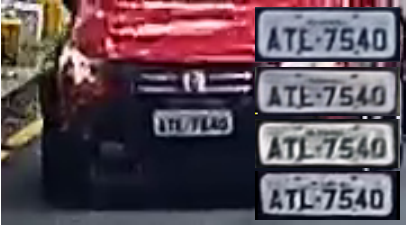}

    \vspace{-2.3mm}

    \caption{\review{Super-resolved \glspl*{lp} generated by our method and baselines from real-world images. The background image shows the original scene from which the \gls*{lr} image was extracted. From top to bottom: \gls*{lp} reconstructions by SR3~\cite{saharia2023image}, PLNET~\cite{nascimento2023super}, our method, and a reference \gls*{hr} image from a different~frame.}}
    \label{fig:real_images_train_test}
\end{figure}

\review{We also explored a dataset of $3,723$ LR-HR image pairs collected from real-world settings.
We allocated $80$\% of the pairs for training, $10$\% for validation, and $10$\% for testing.
\cref{tab:ExperimentalResultsReal} shows the recognition rates achieved on the test images.
The results reinforce the superiority of our super-resolution method, with the \gls*{ocr} model achieving a recognition rate of $39.5$\%, compared to $36.3$\% for PLNET~\cite{nascimento2023super} and $31.7$\% for SR3~\cite{saharia2023image}.
\cref{fig:real_images_train_test} visually demonstrates the effectiveness of the proposed method.
As one example, in the left image, both baseline methods incorrectly reconstructed an ``O'' instead of a ``U,'' with PLNET also introducing noticeable artifacts.
In contrast, our method super-resolved the characters~accurately.}

\begin{table}[!htb]
\centering
\setlength{\tabcolsep}{5pt}
\caption{Recognition rates achieved on real-world images.}

\vspace{-2.25mm}

\review{
\resizebox{0.725\linewidth}{!}{
\begin{tabular}{lccc}
\toprule
\multirow{2}{*}{Test Images} & \multicolumn{3}{c}{\# Correct Characters} \\[-1.25pt] \cmidrule(l{11pt}r{9pt}){2-4} \\[-10pt]
                              & All          & $\geq6$          & $\geq5$          \\
\midrule                
\gls*{hr} (original images) & $90.6$\% & $98.7$\% & $100$\% \\
\gls*{lr} (degraded images)    &  $\phantom{0}9.9$\% & $28.0$\% & $56.2$\% \\
\midrule
LR + SR (SR3~\cite{saharia2023image})  & $31.7$\% & $63.7$\% & $80.1$\% \\
LR + SR (\gls*{plnet}~\cite{nascimento2023super}) &  $36.3$\% & $67.2$\% & $82.5$\% \\
LR + SR (\textbf{Proposed})  & $\textbf{39{.}5}$\textbf{\%} & $\textbf{70.2}$\textbf{\%} & $\textbf{83.1}$\textbf{\%} \\
\bottomrule
\end{tabular} \,
}
}
\label{tab:ExperimentalResultsReal}
\end{table}

\section{Conclusions}
\label{sec:conclusions}

\glsreset{lcofl}

This article proposes a specialized super-resolution method designed to improve the readability of characters and enhance recognition rates in \gls*{lpr} applications.
Our approach involves the implementation of the \gls{lcofl} to guide the network in accurately reconstructing characters according to the \gls{lp} layout, effectively mitigating confusion between structurally similar characters.
Additionally, we enhanced the \gls*{pltfam} model~\cite{nascimento2023super} by introducing shared weights and integrating deformable convolutions, leading to improved feature~extraction.

Our experiments were conducted on the diverse \rodosolalpr~\cite{laroca2022cross} dataset.
The results revealed significantly improved recognition rates in images reconstructed using the proposed method compared to state-of-the-art approaches.
Remarkably, our method led to a recognition rate of $49.8$\% being achieved by the \gls*{ocr} model, whereas the methods proposed in \cite{saharia2023image} and \cite{nascimento2023super} led to recognition rates of $43.1$\% and $39.0$\%, respectively.

The dataset and source code for all experiments are publicly available at~\supplementary.

\review{While our experiments were limited to Brazilian and Mercosur \glspl*{lp}, the findings provide a foundation for future research to explore the method's efficacy across different \gls*{lp} layouts.}

\iffinal
    \section*{\uppercase{Acknowledgments}}

\iffinal
    This study was financed in part by the \textit{Coordenação de Aperfeiçoamento de Pessoal de Nível Superior - Brasil~(CAPES)} - Finance Code 001, and in part by the \textit{Conselho Nacional de Desenvolvimento Científico e Tecnológico~(CNPq)} (\#~315409/2023-1 and \#~312565/2023-2).
    We thank the support of NVIDIA Corporation with the donation of the Quadro RTX $8000$ GPU used for this research.
\else
    The acknowledgments are hidden for review.
\fi

\else
\fi

\bibliographystyle{IEEEtran}

\ifarxiv
\else
    {\footnotesize \bibliography{bibliography-short}}
\fi

\end{document}